\documentclass[journal]{IEEEtran}
\ifCLASSINFOpdf
\else
\fi

\usepackage{amssymb}
\usepackage[colorlinks=False,linkcolor=blue]{hyperref}
\usepackage{amsmath}
\usepackage{graphicx}
\usepackage{caption}
\usepackage[dvipsnames]{xcolor}
\usepackage{subcaption}
\graphicspath{ {images/} }
\usepackage{float}
\usepackage{amsthm}
\usepackage{listings}
\usepackage{algorithm,algcompatible}
\makeatletter
\def\munderbar#1{\underline{\sbox\tw@{$#1$}\dp\tw@\z@\box\tw@}}
\makeatother
\newcommand{\norm}[1]{\left\lVert#1\right\rVert}
\usepackage{algorithm}
\usepackage{algorithmicx}
\usepackage{algpseudocode}
\usepackage{makecell}
\usepackage{booktabs}       
\usepackage{url}
\usepackage{multirow}
\usepackage{lipsum}
\usepackage{bbm}
\usepackage{soul}

\newtheorem{theorem}{Theorem}

\newtheorem{problem}{Problem}
\newtheorem{remark}{Remark}

\newtheorem{definition}{Definition}

\newtheorem{assumption}{Assumption}
\usepackage{lipsum}  

\newcommand{\R}{\mathbb{R}}
\newcommand{\Z}{\mathbb{Z}}
\newcommand{\x}{\mathbf{x}}

\newcommand{\uu}{\mathbf{u}}
\newcommand{\vv}{\mathbf{v}}
\newcommand{\acc}{\mathbf{a}}
\newcommand{\s}{\mathbf{s}}
\newcommand{\p}{\mathbf{p}}
\newcommand{\sig}{\mathbf{\sigma}}

\begin{document}
%
\title{A Formal Characterization of Black-Box System Safety Performance with Scenario Sampling}
%
%
%

\author{Bowen Weng, Linda Capito,  Umit Ozguner, Keith Redmill
\thanks{Bowen Weng, Linda Capito, Umit Ozguner, and Keith Redmill are with the Department of Electrical and Computer Engineering at The Ohio State University, OH, USA.}
}

\maketitle

\begin{abstract}
A typical scenario-based evaluation framework seeks to characterize a black-box system's safety performance (e.g., failure rate) through repeatedly sampling initialization configurations (scenario sampling) and executing a certain test policy for scenario propagation (scenario testing) with the black-box system involved as the test subject. In this letter, we first present a novel safety evaluation criterion that seeks to characterize the actual operational domain within which the test subject would remain safe indefinitely with high probability. By formulating the black-box testing scenario as a dynamic system, we show that the presented problem is equivalent to finding a certain ``almost" robustly forward invariant set for the given system. 
Second, for an arbitrary scenario testing strategy, we propose a scenario sampling algorithm that is provably asymptotically optimal in obtaining the safe invariant set with arbitrarily high accuracy. 
Moreover, as one considers different testing strategies (e.g., biased sampling of safety-critical cases), we show that the proposed algorithm still converges to the unbiased approximation of the safety characterization outcome if the scenario testing satisfies a certain condition. 
Finally, the effectiveness of the presented scenario sampling algorithms and various theoretical properties are demonstrated in a case study of the safety evaluation of a control barrier function-based mobile robot collision avoidance system.
\end{abstract}


%
\IEEEpeerreviewmaketitle

\section{Introduction}\label{sec:introduction}
\IEEEPARstart{A} typical black-box safety evaluation method seeks to justify the system's ability to avoid unaccepted risk events such as physical collisions and explosions. In this letter, we consider the class of intelligent decision-making and control systems as the black-box, which appears in various robotic applications including autonomous vehicles (AV)~\cite{koopman2016challenges} and unmanned aerial vehicles (UAV)~\cite{moss2020adaptive}. Without access to the underlying models or parameters of the black-box system, one common approach for safety evaluation relies on a data-driven methodology of repeatedly (i) sampling initialization states of testing cases, (ii) executing a certain testing policy, and (iii) observing the resulting outcomes. This is known as the ``scenario sampling" testing method. Given the significant amount of required tests and potentially catastrophic consequences of performing them in real life, the scenario sampling based test is typically performed in high-fidelity computer simulators, and is occasionally complemented with a limited number of trials in a controlled real-world proving grounds.

In general, scenarios are defined in a concrete manner, where the set of testing cases (scenarios) are predetermined and the scenario propagation is often independent of the test subject's response. Every time the black-box test subject is updated, it is expected to go through the same set of concrete testing scenarios. This is commonly observed in computer software regression tests~\cite{rothermel1996analyzing} and the standard testing procedures for Advanced Driving Assistant System (ADAS)~\cite{van2017euro}. However, as the black-box system becomes more advanced, the concrete scenario becomes less effective as it becomes ``hackable" and incomplete~\cite{hauer2020re,capito2020modeled}. An adaptive scenario design becomes thus a natural alternative.

Existing efforts related to the adaptive scenario sampling-based black-box system safety evaluation can be classified into two categories, (i) the falsification-based approaches and (ii) the validation approaches~\cite{riedmaier2020survey}. In general, the falsification-based method is biased towards corner-case generation with safety-critical instances. This is often formulated as a path planning problem where the unacceptable risky event becomes the target state and one seeks to derive a sequence of actions that drives the given initial state to the risky outcome~\cite{nahhal2007test}. As the Operational Design Domain (ODD) has become more complex and the black-box system more sophisticated, some recent work also formulate the falsification problem as a reinforcement learning (RL) problem~\cite{corso2020survey}, and RL solutions are derived correspondingly~\cite{corso2019adaptive, ding2021multimodal}.

On the other hand, the validation approaches seek to derive an overall safety characterization of the black-box system, where the most well-adopted characterization is the failure rate (e.g. crash rate for AVs). This typically requires sufficient coverage over the given ODD through the Monte Carlo sampling technique, which can be sample inefficient~\cite{kalra2016driving}. To deal with this, the ``importance sampling" methodology is introduced with success in AV safety evaluation~\cite{zhao2017accelerated,feng2021intelligent} and has also been extended to other robotic safety validation applications~\cite{moss2021pomdpstresstesting}. The method focuses on sampling ``important" scenarios that are safety-critical, and manages to obtain the failure rate by associating the sampled scenario with its importance function (i.e., exposure rate) to the nominal distribution. Note that such a biased sampling towards important scenarios still requires sampling and executing some obviously safe and normal cases to ensure that the failure rate estimate is unbiased.

In general, neither the falsification assessment nor the failure rate validation is sufficient to characterize the test subject's safety performance profile. For example, it is unclear which states could lead to failure/non-failure events with high probability. To take advantage of the domain knowledge, some ``gray-box" methodologies are introduced~\cite{li2020av}, such as the Backtracking Process Algorithm~\cite{hejase2020methodology}, and temporal logic approaches~\cite{dokhanchi2018evaluating}. Such methods often rely on heuristics, expert knowledge, and the modeling of (part of) the black-box system (i.e., system identification). 

In this letter, we propose a novel black-box system safety characterization framework with scenario sampling by approximating the safe operable domain as a robustly forward invariant set using the data-driven approach. We propose an algorithm that is provably asymptotically optimal in obtaining such an approximation with arbitrarily high accuracy. We further summarize the contributions as follows.
\begin{figure}[!t]
    \centering
    \vspace{2mm}
    \includegraphics[trim={0cm 0 0cm 0.5cm},clip,width=0.45\textwidth]{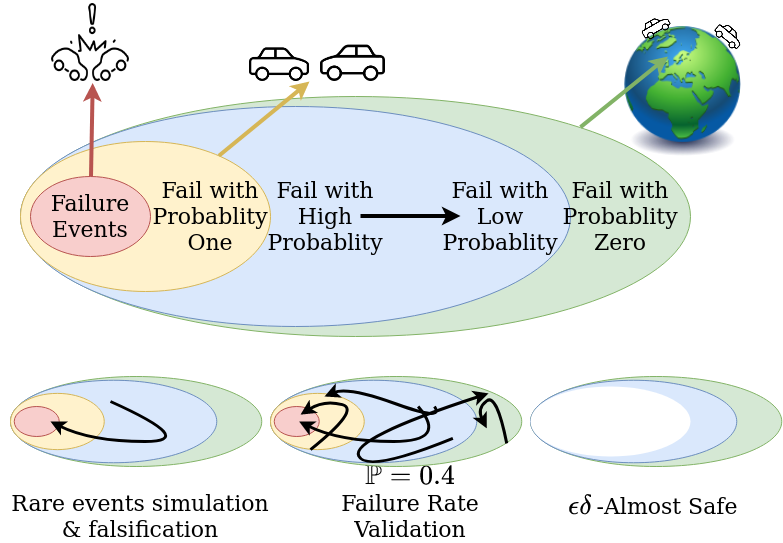}
    \caption{\small{A conceptual comparison among the falsification-based safety validation methodology, the failure rate characterization, and the proposed safety evaluation scheme with data-driven invariant set characterization.}}
    \label{fig:overview}
    \vspace{-5mm}
\end{figure}

\textbf{A novel safety characterization criterion: } We propose a safety characterization that approximates the actual safe operable domain of the black-box system as an ``almost" robustly forward invariant set. This complements the commonly adopted safety characterization that focuses on the unbiased estimate of the failure rate, by considering more detailed safety performance properties. The difference between the proposed criterion and other safety characterization methods is conceptually illustrated in Fig.~\ref{fig:overview}.

\textbf{A formal safety evaluation framework: } For the proposed safety characterization, we further present an algorithm that is provably asymptotically optimal. That is, as the number of sampled scenarios tends to infinity, the probability for the presented algorithm to obtain the actual forward invariant safe operable domain tends to one. 

\textbf{Asymptotic safety characterization consensus: } In general, the ``safety characterization consensus" represents the phenomenon where different scenario testing policies lead to the same safety characterization outcome. This is generally difficult to achieve as, for example, the ``worst-case-only" testing policy will not derive the same estimate of failure rate as that found with Monte Carlo sampling. However, supplied with a rigorously defined condition, we show that the proposed algorithm ensures the asymptotic safety characterization consensus. We also present empirical evidence that supports the derived theoretical property.

\section{Preliminaries and Problem Formulation}\label{sec:prelim&prob}
This section begins by presenting the black-box system testing scenario with the dynamic system. Such a system interpretation further gives rise to the definition and validation of the ``almost" safe set and the proposed safety characterization problem formulation.

\subsection{The black-box system testing scenario}
In general, consider a black-box system that admits the discrete-time nonlinear motion dynamics in the form of
\begin{equation}\label{eq:s0-dyn}
    \s_0(t+1) = f_0(\s_0(t), \uu_0(t); \boldsymbol\omega_0(t)),
\end{equation}
where $t \in \Z$ will be subsequently omitted unless needed. Then $\s_0 \in \mathcal{S}_0 \subseteq \R^{n_0}$ is the state, $\uu_0 \in \mathcal{U}_0\subseteq \R^{m_0}$ is the actions and $\boldsymbol\omega_0 \in \Omega_0 \subseteq \R^{w_0}$ represents the disturbances and uncertainties. Taking such a black-box system as the test subject, one then involves other participants and controlled features to form a testing scenario. The black-box system admits a certain feedback control policy based on a perceivable subset of state observations as the input, i.e.,
\begin{equation}\label{eq:s0-ctrl}
    \uu_0 = \pi_0\left(\phi(\s_0, \s_s; \boldsymbol\omega_{\phi}); \boldsymbol\omega_{\pi_0}\right), 
\end{equation}
where $\s_s \in \mathcal{S}_s \subseteq \R^{n_s}$ represents all non-test-subject states. The mapping $\phi: \mathcal{S}_0\times\mathcal{S}_s\times\Omega_{\phi} \rightarrow \mathcal{S}_p\subseteq\R^{p}$ is a perception function that takes a subset or a projected subset of observed states in the neighborhood of the test subject. With the uncertainties $\boldsymbol\omega_{\pi_0}\in\Omega_{\pi_0}$, we have the test subject policy $\pi_0: \mathcal{S}_p \times \Omega_{\pi_0} \rightarrow \mathcal{U}_0$. Moreover, the state of all non-test-subject actors also admits certain motion dynamics, in the form of
\begin{equation}\label{eq:ss-dyn}
    \s_s(t+1) = f_s(\s_s(t), \uu_s(t); \boldsymbol\omega_s(t)).
\end{equation}
where the action $\uu_s \in \mathcal{U}_s \subseteq \R^{m_s}$ and the uncertainties $\boldsymbol\omega_s\in\Omega_s$. 
Note that both the dynamics transition $f_0$ in~\eqref{eq:s0-dyn} and the control policy for the test subject \eqref{eq:s0-ctrl} remain unknown, hence the name ``black-box". Integrating \eqref{eq:s0-dyn} with \eqref{eq:ss-dyn}, we have the scenario dynamics in the expanded form as
\begin{equation}\label{eq:s-dyn}
    \begin{aligned}
    & \s(t+1) = \begin{bmatrix}\s_0(t+1) \\ \s_s(t+1)\end{bmatrix} \\ & = \begin{bmatrix}f_0(\s_0(t), \pi_0\left(\phi(\s_0(t), \s_s(t); \boldsymbol\omega_{\phi}(t)); \boldsymbol\omega_{\pi_0}(t)\right); \boldsymbol\omega_0(t)) \\ f_s(\ \s_s(t),\ \uu_s(t);\ \boldsymbol\omega_s(t)\ )\end{bmatrix}\\& = \hat{f}(\s(t), \uu_s(t); \hat{\boldsymbol\omega}(t))
    \end{aligned}
\end{equation}
The safety evaluation with scenario sampling thus relies on (i) providing an initialization state $\s(0)=\s^0$ for the test subject and all other participants, and (ii) executing a (centralized) feedback control policy 
\begin{equation}\label{eq:s-ctrl}
    \uu_s=\pi_s(\s;\boldsymbol\omega_{\pi_s}),
\end{equation}
for the run of a scenario. Note that with the uncertainties $\boldsymbol\omega_{\pi_s}\in\Omega_{\pi_s}$, the policy \eqref{eq:s-ctrl} is essentially stochastic. Another equivalent configuration is to formulate~\eqref{eq:s-ctrl} as a Markov Decision Process~\cite{corso2019adaptive} or an automaton~\cite{capito2020modeled}. We are now ready to present the formal definition of a black-box system testing scenario.
\begin{definition}{[\textbf{Black-Box System Testing Scenario}]}
    A black-box system testing scenario is a dynamic system that admits the motion of
    \begin{equation}\label{eq:scenario-dyn}
        \s(t+1)\!=\!\hat{f}\!\left(\s(t),\!\pi_s(\s(t);\!\boldsymbol\omega_{\pi_s}(t))\!; \!\hat{\boldsymbol\omega}(t)\!\right)\!=\!f(\s(t)\!;\!\boldsymbol\omega(t)),
    \end{equation}
    with state $\s \in \mathcal{S} = \mathcal{S}_0 \times \mathcal{S}_s$ and the composed uncertainties $\boldsymbol\omega\in\Omega$.
\end{definition}
Present an initial condition $\s(0)=\s^0$ and a scenario testing policy in the form of~\eqref{eq:s-ctrl}, we then have that a run of a scenario is $\{\s(t)\}_{t=0,\ldots,K}$. It is a finite-step run of scenario for $K<\infty$ and an infinite-step run of scenario if $K=\infty$. In this letter, we consider a run of scenario of finite steps and $K\geq2$ for practical applicability, and the obtained property generalizes to $K=\infty$ as well.
\begin{figure}
    \centering
    \vspace{2mm}
    \includegraphics[width=0.45\textwidth]{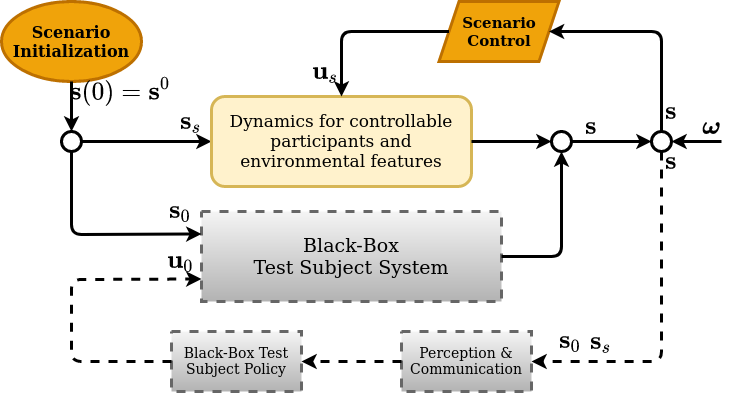}
    \caption{\small{Interpreting the black-box system testing scenario as a dynamic system.}}
    \label{fig:black-box-scenario}
    \vspace{-5mm}
\end{figure}

\subsection{The ``almost-safety": definition and validation}
Let $\mathcal{C}\subseteq\mathcal{S}$ be the set of states deemed with unaccepted risk, such as collisions. The definition of safety is thus naturally obtained from a set invariance perspective.
\begin{definition}\label{def:safety}
    The test subject is safe in $\Sigma \subseteq \mathcal{S}\setminus\mathcal{C}$ for \eqref{eq:scenario-dyn} if and only if $\forall \s(0) \in \Sigma, \forall \boldsymbol\omega \in \Omega$, $\s(t) \in \Sigma, \forall t > 0$. That is, the test subject is safe in $\Sigma$ if $\Sigma$ is robustly forward invariant for~\eqref{eq:scenario-dyn}.
\end{definition}
In the practice of the data-driven methodologies, the absolute safety defined as above is of very little practical value as real-world safety evaluation is primarily interested in safety assurance up to a sufficiently high level (e.g., sufficiently small fatality rate~\cite{kalra2016driving} and sufficiently low risk~\cite{stolte2017hazard}). We then, inspired by~\cite{wang2020scenario, weng2021towards}, introduce the idea of having a black-box system being safe for ``almost" everywhere except for an arbitrarily small subset. We start from the ``extended $\delta$-covering set", which is a special discretization scheme. 
\begin{definition}{[\textbf{Extended $\delta$-Covering Set}]}\label{def:ext-delta-cover}
    Given $ \mathbf{x} \in \mathcal{X} \subset \R^x$ and $\boldsymbol\delta \in \R_{>0}^x$, let $\mathcal{N}_{\delta}(\mathbf{x})$ be the ``extended $\delta$-neighbourhood" of $\mathbf{x}$, i.e., 
    \begin{equation}\label{eq:ext-delta-neib}
        \forall \mathbf{x}'\in \mathcal{N}_{\delta}(\mathbf{x}), |\mathbf{x}-\mathbf{x}'| \leq \boldsymbol\delta.
    \end{equation}
    We claim that $\Phi_{\delta}^{\mathcal{X}}$ is an extended $\delta$-covering set of $\mathcal{X}$ if for some $k \in \Z$ and $\mathbf{x}_i\in\mathcal{X}, i=1,\ldots,k$, we have
    \begin{equation}
        \Phi_{\delta}^{\mathcal{X}}\!=\!\bigcup_{i\in\{1,\ldots,k\}} \mathcal{N}_{\delta}(\mathbf{x}_i) \supseteq \mathcal{X} \text{ and } \Phi_{x}^{\mathcal{X}}\!=\!\{\mathbf{x}_i\}_{i\in\{1,\ldots,k\}} \subseteq \mathcal{X}.
    \end{equation}
    Furthermore, $\Phi_{x}^{\mathcal{X}}$ are centroids of $\Phi_{\delta}^{\mathcal{X}}$.
\end{definition}
Note that the inequality in \eqref{eq:ext-delta-neib} is element-wise, i.e., $|\x|\leq\boldsymbol\delta$ implies $|\x^i|\leq\boldsymbol\delta^i, \forall i\in \Z_{x}$ where $\x^i$ and $\boldsymbol\delta^i$ indicate the $i$-th element of $\x$ and $\boldsymbol\delta$, respectively. Comparing with the previous covering set definition~\cite{weng2021towards}, Definition~\ref{def:ext-delta-cover} enables a covering set that considers the system of unbalanced states (e.g., a state space with both wide-range positioning states and periodic angular states defined over a small interval in $\R$). In practice, the extended $\delta$-covering set for $\Sigma$ can be built in two ways, (i) a grid-based decomposition of the state space where all centroids are manually selected, or (ii) an ``experience replay" manner where one takes points from randomly collected runs of scenarios to ensure the desired coverage. Note that the second approach also ``connects" some of the centroids through the dynamics~\eqref{eq:scenario-dyn}, thus $\Phi_{\sigma}^{\mathcal{X}}$ can also be denoted as a $\delta$-disk graph as
\begin{equation}
    G(\Phi_{\sigma}^{\mathcal{X}}, E_{\sigma}^{\mathcal{X}}),
\end{equation}
with the vertices $\Phi_{\sigma}^{\mathcal{X}}$ and the edges $E_{\sigma}^{\mathcal{X}}$. It is also immediate that given $\mathcal{X}$, the corresponding extended $\delta$-covering set is not necessarily unique, and $\lim_{\boldsymbol\delta\rightarrow0}\Phi_{\delta}^{\mathcal{X}}=\mathcal{X}$. The notion of ``almost" safe is then formally presented through the following definition.
\begin{definition}{[\textbf{$\epsilon\delta$-Almost Safe}]}\label{def:almost-safe}
    Consider $\mathcal{S}\subseteq \R^n$, $\boldsymbol\delta \in \R_{>0}^{n_s}$ and $\epsilon\in\R_{>0}$. The test subject is $\epsilon\delta$-almost safe in $\Sigma \subseteq \mathcal{S}\setminus\mathcal{C}$ for \eqref{eq:scenario-dyn} if there exists an extended $\delta$-covering set of $\Sigma$, $\Phi_{\delta}^{\Sigma}$ and its corresponding centroids $\Phi_{\sigma}^{\Sigma}$, such that for all $\boldsymbol\omega_{\pi_s} \in \Omega_{\pi_s}, \boldsymbol\omega \in \Omega$, 
    \begin{equation}\label{eq:almost-safe}
        \mathbb{P}\Big(\big\{ \boldsymbol\sig \in \Phi_{\sigma}^{\Sigma} : f\left(\boldsymbol\sigma, \pi_s(\boldsymbol\sigma;\boldsymbol\omega_{\pi_s}); \boldsymbol\omega\right) \not\in \Phi_{\delta}^{\Sigma}\big\}\Big)\leq \epsilon.
    \end{equation}
\end{definition}
Finally, in order to claim that given set is an $\epsilon\delta$-almost safe set, one would require observing a sufficient amount of ``safe" sampled runs of scenarios. Such a sampling sufficiency is formally defined through the following theorem proven in~\cite{wang2020scenario,weng2021towards}.
\begin{theorem}{[\textbf{$\epsilon\delta$-Almost Safety Validation}]}\label{thm:finite-sample-almost-safe}
    Consider $\mathcal{S}\subseteq \R^n$, $\boldsymbol\delta \in \R_{>0}^{n_s}$ and $\epsilon\in\R_{>0}$. Let $\Phi_{\delta}^{\Sigma}$ be an extended $\delta$-covering set of $\Sigma \subseteq \mathcal{S}$. Consider $N$ runs of scenarios with $K$ steps ($K \geq 2$), leading to the collected set of states as $\mathcal{R} = \{\s_i\}_{i=1,\ldots,NK}$.
    Let the initial set of states $\{\s_i^0\}_{i=1,\ldots,N}$ be i.i.d. w.r.t. the underlying distribution on $\Phi_{\sigma}^{\Sigma}$. The test subject is $\epsilon\delta$-almost safe in $\Sigma$ with probability no smaller than $1\!-\!\beta$ if
    \begin{equation}
        \mathcal{R} \cap \mathcal{C} = \emptyset \text{ and } N \geq \frac{\ln{\beta}}{\ln{(1-\epsilon)}}.
    \end{equation}
\end{theorem}

\subsection{The optimal safety characterization problem} 
In general, the designer specified ODD of an engineering product is not necessarily the true ODD within which the product will remain safe persistently. The scenario sampling approach is thus a natural solution to identify the real ODD in a data-driven manner. Formally speaking, the scenario sampling algorithm $\mathcal{ALG}$ for safety characterization considered in this letter is defined as
\begin{equation}\label{eq:alg}
    \mathcal{ALG}_{\pi_s}(N, \mathcal{S}^0), \mathcal{ALG}_{\pi_s}: \Z \times \mathcal{S} \rightarrow \mathcal{S}.
\end{equation}
When presented with the initialization set $\mathcal{S}^0$ (e.g.,$\mathcal{S}^0=\mathcal{S}\setminus\mathcal{C}$) and the feedback control testing policy $\pi_s(\cdot)$ as defined in~\eqref{eq:s-ctrl}, the algorithm seeks to determine the actual safe subset $\Sigma \subseteq \mathcal{S}^0$ by sampling $N$ runs of a scenario. Note that the actual safe domain is not necessarily unique. Throughout this study, we are interested in finding the particular safe state set with the maximum cardinality, or equivalently, we seek to find the tightest sup-set of all safe subsets, which presents an extra challenge to the algorithm design. Before introducing the method, let us formally define the optimal safety characterization problem as follows.
\begin{problem}{[\textbf{The Optimal Safety Characterization Problem}]}\label{prob}
    Consider a black-box system of dynamics~\eqref{eq:s0-dyn} operating in the state space $\mathcal{S}\subseteq \R^n$. Suppose that the optimal safe subset $\mathcal{S}^* \subseteq \mathcal{S}\setminus\mathcal{C}$ exists and has cardinality $c^*$. Let $c(\Sigma), c:\mathcal{S} \rightarrow \R_{\leq0}$ be the cardinality of the set $\Sigma \subseteq \mathcal{S}$. Given an initial set $\mathcal{S}^0 \subseteq \mathcal{S}\setminus\mathcal{C}$, the optimal safety characterization problem seeks to find an asymptotically optimal scenario sampling algorithm $\mathcal{ALG}_{\pi_s}(N, \mathcal{S}^0)$ in the form of~\eqref{eq:alg} such that
    \begin{equation}
        \limsup_{N\rightarrow\infty}{\mathbb{P}\Big(\big\{c(\mathcal{ALG}_{\pi_s}(N, \mathcal{S}^0))=c^*\big\}\Big)=1}
    \end{equation}
\end{problem}
That is, as one samples more runs of scenarios, the probability of obtaining the optimal safe set tends to one. Moreover, the following remark reveals an intrinsic property that connects the desired outcome of Problem~\ref{prob} with the classic failure rate characterization.
\begin{remark}\label{rmk:equal_to_fr}
    Comparing the cardinality of the optimal safe set $|\mathcal{S}^*|$ with the cardinality of the complete non-failure set $|\mathcal{S}\setminus \mathcal{C}|$, we immediately obtain the failure rate as $1-\frac{|\mathcal{S}^*|}{|\mathcal{S}\setminus \mathcal{C}|}$. 
\end{remark}
The above remark also reveals a fundamental comparison between the safety characterizations through the failure rate and the proposed method from the statistician's perspective. That is, the failure rate estimate is from the ``frequentist" inference, where the probability estimate is inferred through infinite sampling of runs of scenarios. On the other hand, the idea of the almost safe set corresponds to the ``Bayesian" view of probability as $\epsilon$ implies the prior probability, and the scenario sampling is simply a way of obtaining the posterior probability within a certain confidence interval. Such different views of probability will not affect the final safety evaluation outcome (as discussed in the above remark), but may lead to algorithms of different properties, such as sampling efficiency as we will demonstrate in Fig.~\ref{fig:cbf_sample_efficiency} later.

Note that Problem~\ref{prob} is in general challenging with some seemingly working but theoretically invalid solutions as discussed in~\cite{weng2021towards}.

Finally, we conclude this section with the following remark emphasizing the scenario-based nature of the testing approach shared by many related work in the literature~\cite{li2020av,feng2021intelligent,capito2020modeled}.
\begin{remark}\label{rmk:scenario-based}
    Throughout the letter, we are primarily interested in the scenario-based testing methodology~\cite{riedmaier2020survey}, which typically occurs in computer simulators and self-contained test proofing grounds. As a result, we further assume that all traffic participants specified by the operable domain are controllable, and the states of all non-specified participants are considered as uncertainties and disturbances characterized by $\boldsymbol\omega$ in~\eqref{eq:scenario-dyn}. All states of interest specified by the domain definition are fully observable. Finally, the test subject policy $\pi_0(\cdot)$ remains stable throughout the testing process.
\end{remark}

\section{Main Method}
Our proposed solution to solve Problem~\ref{prob} is given by Algorithm~\ref{alg:alg-overview} considering Definition~\ref{def:almost-safe}.
\begin{algorithm}[b]
    \begin{algorithmic}[1]
    \Require $N, \mathcal{S}^0 (\mathcal{S}^0\subseteq \mathcal{S}\setminus\mathcal{C})$, scenario control $\pi_s(\cdot)$ in \eqref{eq:s-ctrl}, decay coefficients $\lambda_{\epsilon}, \lambda_{\delta}, \lambda_{\beta} \in (0,1)$, $\bar{\mathcal{S}}^0 \subseteq \mathcal{S}\setminus\mathcal{S}^0$.
    \State {\bf Initialize}: $\epsilon, \boldsymbol\delta, \beta, i=1$.
    \State {\bf While} $i<N$:
    \State {\ \ \ \ {\bf $\epsilon\delta$-Almost Safe Set Quantification}: Find $\epsilon\delta$-almost safe set $\Phi_{\delta}^{\Sigma}$ such that $\Sigma=\mathcal{S}^*$ with confidence level at least $(1-\beta)$ and update $\bar{\mathcal{S}}^0$ with $N_{\sigma}$ runs of scenarios}
    \State {\ \ \ \ $\boldsymbol\delta'\leftarrow\lambda_{\delta}\boldsymbol\delta$}
    \State {\ \ \ \ $\mathcal{S}^0=\Phi_{\delta'}^{\Phi_{\delta}^{\Sigma}}$}
    \State {\ \ \ \ $\boldsymbol\delta\leftarrow\boldsymbol\delta', \epsilon \leftarrow \lambda_{\epsilon}\epsilon, \beta\leftarrow\lambda_{\beta}\beta$}
    \State {\ \ \ \ $i \leftarrow i+N_{\sigma}$}
    \Ensure $\lim_{N\rightarrow\infty} \Phi_{\delta}^{\Sigma}=\lim_{N\rightarrow\infty}\Phi_{\sigma}^{\mathcal{S}^*}=\mathcal{S}^*$
    \end{algorithmic}
    \caption{Scenario sampling with $\epsilon\delta$-decay $\mathcal{ALG}^{\epsilon\delta}_{\pi_s}(N, \mathcal{S}^0)$} \label{alg:alg-overview}
\end{algorithm}
The key step in Algorithm~\ref{alg:alg-overview} is in line 3 where one is expected to find a $\epsilon\delta$-almost safe set with at least $(1-\beta)$ confidence level in a finite number of sampled runs of scenarios. If such an expectation is satisfied, with $\lambda_{\epsilon}, \lambda_{\delta}, \lambda_{\beta} \in (0,1)$, it follows that Algorithm~\ref{alg:alg-overview} is asymptotically optimal in solving Problem~\ref{prob}. The algorithm referenced at line 3 is referred to as the $\epsilon\delta$-almost safe set quantification. The remainder of this section will primarily focus on (i) the critical set of states (``where" the quantification should be performed), and (ii) the state pruning and exploration with scenario sampling (``how" the quantification should be performed). We will also discuss different choices of scenario testing policy $\pi_s(\cdot)$ and how they affect the performance of the proposed algorithm.

\subsection{Critical state set}
We start with the critical set of states, which shows that the $\epsilon\delta$-almost safe property only requires investigating a subset of states. This significantly reduces the computational burden of the proposed algorithm. Consider the following assumption.
\begin{assumption}\label{asp:bound_s}
    Consider the dynamics~\eqref{eq:scenario-dyn}, assume $\exists \bar{\s} > 0$ such that $\forall \boldsymbol\omega\in\Omega$, $\norm{f(\s; \boldsymbol\omega)-\s}_{\infty} \leq \bar{\s}, \forall \s \in \mathcal{S}$.
\end{assumption}
The above property is a direct outcome of $f$ being Lipschitz continuous and is also valid in practice. For some practical systems that are not necessarily Lipschitz continuous (e.g. the locally Lipschitz system and the hybrid system), the above property may still hold. Let $\partial\mathcal{S}$ be the ``boundary" of the compact set $\mathcal{S}$.

\begin{figure}[!t]
    \centering
    \includegraphics[width=0.45\textwidth]{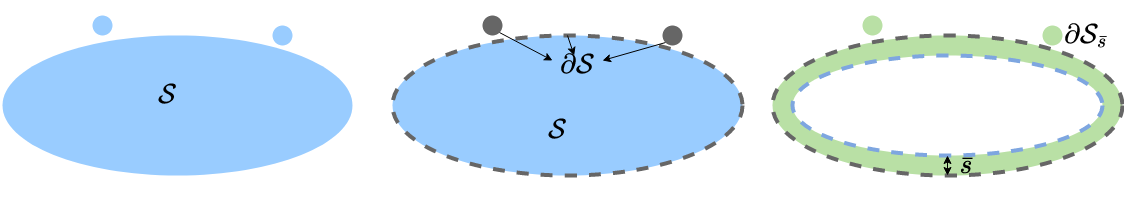}
    \caption{\small{A conceptual example of the set boundary and the critical state set. Note that $\mathcal{S}$ consists of a continuous set and two extra points. Those two points are also part of $\partial\mathcal{S}$ by definition.}}
    \label{fig:bss}
    \vspace{-5mm}
\end{figure}
Note that the boundary of a set is connected to both the interior and the exterior of the set (as shown in Fig.~\ref{fig:bss}). The following definition denotes a specific neighbourhood region near the boundary. 
\begin{definition}{[\textbf{$\bar{s}$-Critical State Set}]}\label{def:boundary-set}
    Let $\bar{s}$ defined as in Assumption~\ref{asp:bound_s}. $\partial\mathcal{S}_{\bar{s}}$ is the $\bar{s}$-critical state set of $\mathcal{S}$ if
    \begin{equation}
        \partial\mathcal{S}\!\subseteq\!\partial\mathcal{S}_{\bar{s}}\!\subseteq\!\mathcal{S} \text{ and } \forall\!\s_i\!\in\!\partial\mathcal{S}_{\bar{s}}, \s_j\!\in\! \partial\mathcal{S}_{\bar{s}}, \norm{\s_i-\s_j}_{\infty}\!\leq\!\bar{\s}. 
    \end{equation}
\end{definition}

Combining Assumption~\ref{asp:bound_s} and Definition~\ref{def:boundary-set}, we have the following theorem that gives an equivalent condition to the safe set defined in Definition~\ref{def:safety}.
\begin{theorem}\label{thm:bss}
    The test subject is safe in $\Sigma \subseteq \mathcal{S}\setminus\mathcal{C}$ for \eqref{eq:scenario-dyn} if and only if the $\bar{s}$-critical state set of $\Sigma$, $\partial\Sigma_{\bar{s}} \subseteq \Sigma$ by Definition~\ref{def:boundary-set}, satisfies
    \begin{equation}
        \forall \boldsymbol\omega \in \Omega, \forall \s \in \partial\Sigma_{\bar{s}}, f(\s;\boldsymbol\omega) \in \Sigma.
    \end{equation}
\end{theorem}
That is, to investigate the safety property of the test subject in the set $\Sigma$, one is only required evaluate the $\bar{s}$-critical state set of $\Sigma$. The proof of Theorem~\ref{thm:bss} can readily be adapted from~\cite{weng2021towards} and is omitted. In the remainder of this letter, the set $\Sigma$ and the $\bar{s}$-critical state set of $\Sigma$ are often interchangeable, given the equivalence implied by Theorem~\ref{thm:bss} in terms of safety property evaluation.
\subsection{State pruning, exploration, and almost-safety quantification }
\begin{figure}[!t]
    \centering
    \vspace{1mm}
    \includegraphics[width=0.49\textwidth]{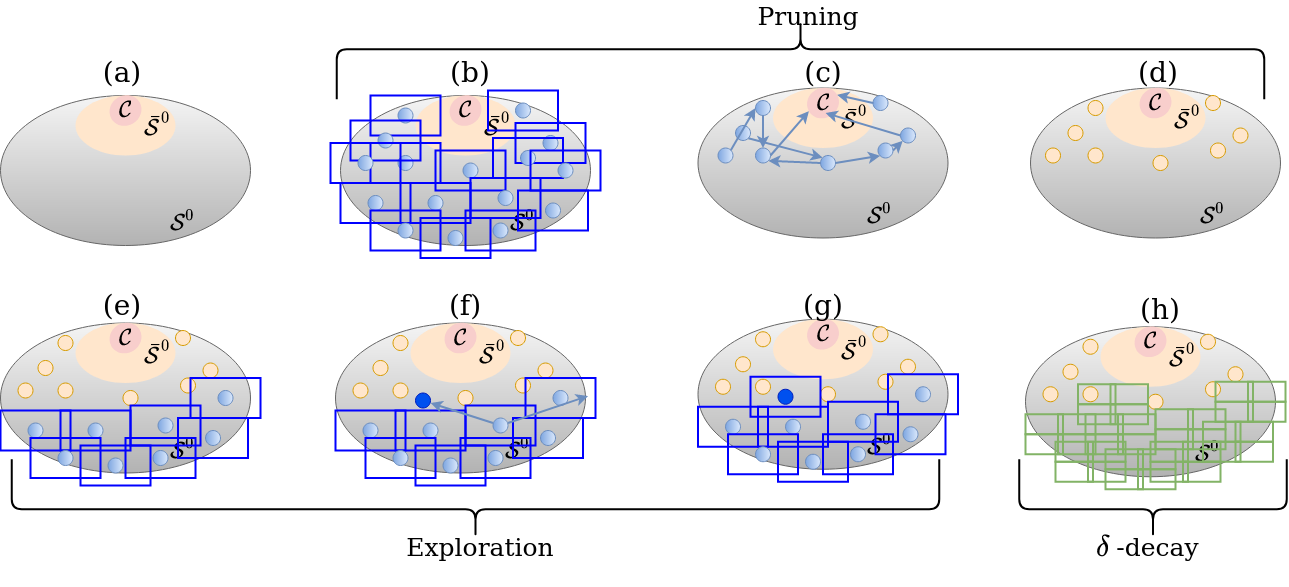}
    \caption{\small{A simple example of the key procedures, pruning and exploration, in the proposed $\epsilon\delta$-almost safe set quantification algorithm (Algorithm~\ref{alg:quant}), and $\delta$-decay described in Algorithm~\ref{alg:alg-overview} (line 5). The round dots and the colored rectangles denote the centroids and the $\delta$-neighbourhood of the centroids, respectively}}
    \label{fig:pruning_exploration}
    \vspace{-6mm}
\end{figure}
In general, given $\epsilon, \boldsymbol\delta, \beta$, and the initial $\mathcal{S}^0$, the proposed $\epsilon\delta$-almost safe set quantification algorithm starts with a $\delta$-covering set of $\mathcal{S}^0$, $\Phi_{\delta}^{\Sigma}$ ($\Sigma=\mathcal{S}^0$) (or equivalently by Theorem~\ref{thm:bss} $\Phi_{\delta}^{\Sigma_{\bar{s}}}$ if supplied with an appropriate $\bar{s}$), as shown in Fig.~\ref{fig:pruning_exploration} (a) and (b). One then removes states that propagate motion trajectories to $\mathcal{C}$ from the instantaneous $\Phi_{\delta}^{\Sigma}$ (i.e., pruning as shown in Fig.~\ref{fig:pruning_exploration} (c) and (d)), and expands $\Phi_{\delta}^{\Sigma}$ with states that do not belong to $\bar{\mathcal{S}}^0$ or $\mathcal{C}$ (i.e., exploration as shown in Fig.~\ref{fig:pruning_exploration} (e) to (g)). The pruning and exploration procedures are performed synchronously until no more states are added to or removed from $\Phi_{\delta}^{\Sigma}$ for a sufficient number of sampled runs of scenarios to claim the almost-safe property with the desired confidence level of $1-\beta$ by Theorem~\ref{thm:finite-sample-almost-safe}. 

\begin{algorithm}[H]
    \begin{algorithmic}[1]
    \Require $\epsilon, \boldsymbol\delta, \beta, \mathcal{S}^0, \bar{\mathcal{S}}^0, K, \mathcal{C}, \bar{s}$
    \State {Let $\Phi_{\sigma}^{\partial\Sigma_{\bar{s}}} \subseteq \partial\Sigma_{\bar{s}} = \partial\mathcal{S}^0_{\bar{s}} \subseteq \Phi_{\delta}^{\partial\Sigma_{\bar{s}}}$ and $G(\Phi_{\sigma}^{\partial\Sigma_{\bar{s}}}, E_{\sigma}^{\partial\Sigma_{\bar{s}}})$}
    \State {Let $N = \frac{\ln{\beta}}{\ln{(1-\epsilon)}}, i=1$}
    \State {{\bf While} $i<N$:}
    \State {\ \ \ \ i.i.d. sample of $\s^0 \in \Phi_{\sigma}^{\Sigma}$}
    \State {\ \ \ \ collect a run of scenario, $\{\s_j\}_{j=1,\ldots,K}$ initialized at $\s^0$}
    \State {\ \ \ \ {\bf For }$j=1,\ldots,K$ {\bf do}}
    \State {\ \ \ \ \ \ \ \ {\bf If} $\s_j \in \mathcal{C}$:}
    \State {\ \ \ \ \ \ \ \ \ \ \ \ $\Phi_{\sigma}^{\partial\Sigma_{\bar{s}}}$.remove$(reachable(\s^0, G(\Phi_{\sigma}^{\partial\Sigma_{\bar{s}}}, E_{\sigma}^{\partial\Sigma_{\bar{s}}})))$}
    \State {\ \ \ \ \ \ \ \ \ \ \ \ $\bar{\mathcal{S}}^0$.add$(reachable(\s^0, G(\Phi_{\sigma}^{\partial\Sigma_{\bar{s}}}, E_{\sigma}^{\partial\Sigma_{\bar{s}}})))$}
    \State {\ \ \ \ \ \ \ \ \ \ \ \ $i=1$}
    \State {\ \ \ \ \ \ \ \ {\bf Else If} $\s_j \notin \Phi_{\delta}^{\partial\Sigma_{\bar{s}}}$:}
    \State {\ \ \ \ \ \ \ \ \ \ \ \  $E_{\sigma}^{\partial\Sigma_{\bar{s}}}$.append$((\s^0,\s_j))$}
    \State {\ \ \ \ \ \ \ \ \ \ \ \ $i=1$}
    \State {\ \ \ \ \ \ \ \ {\bf End If}}
    \State {\ \ \ \ {\bf End For}}
    \State {\ \ \ \ $i\leftarrow i+1$}
    \Ensure $\Phi_{\delta}^{\partial\Sigma_{\bar{s}}}$, (optional: $\bar{\mathcal{S}}^0, G(\Phi_{\sigma}^{\partial\Sigma_{\bar{s}}}, E_{\sigma}^{\partial\Sigma_{\bar{s}}})$)
    \end{algorithmic}
    \caption{$\epsilon\delta$-Almost Safe Set Quantification} \label{alg:quant}
\end{algorithm}

\begin{theorem}\label{thm:opt-alg}
    Algorithm~\ref{alg:alg-overview} with the $\epsilon\delta$-almost safe set quantification step executed as specified by Algorithm~\ref{alg:quant} is asymptotically optimal.
\end{theorem}
Let \textit{reachable}$(\s, G(\cdot,\cdot))$ be all vertices that directly or indirectly connect to $\s$ on the graph. We have the above description formally summarized in Algorithm~\ref{alg:quant}. Note that $\text{.remove}$ and $\text{.add}$ are notionally methods for the class of set operations. The conceptual illustration of the pruning (line 7-10) and the exploration (line 11-13) are also presented in Fig.~\ref{fig:pruning_exploration}. We also have Theorem~\ref{thm:opt-alg} presented with the sketch of proof.
The proof sketch for Theorem~\ref{thm:opt-alg} is in three steps. First, consider Algorithm~\ref{alg:quant} with fixed $\epsilon, \boldsymbol\delta$, and $\beta$, one can prove the algorithm is probabilistically complete as the process always converges to a stationary set as the number of sampled runs of scenarios tends to infinity. Second, one can show that such a stationary set is the optimal set specified in Problem~\ref{prob} through contradictions. Finally, considering the decaying mechanism introduced in Algorithm~\ref{alg:alg-overview}, the asymptotic optimal property can be directly obtained with all decaying coefficients $\lambda_{\epsilon}, \lambda_{\delta}, \lambda_{\beta} \in (0,1)$.

In practice, Algorithm~\ref{alg:alg-overview} does not run indefinitely, and the termination is determined with sufficiently appropriate $\epsilon, \boldsymbol\delta$, and $\beta$, as we show in Section~\ref{sec:case}.
\subsection{Towards the safety characterization consensus with different scenario testing policies}
Up to this point, all of the analyses have been based upon a pre-determined scenario testing policy $\pi_s(\cdot)$. In practice, the testing policy is not necessarily unique. But it is expected that the final safety characterization remains persistent regardless of the selected scenario testing policy. This property is referred to as the ``safety characterization consensus".

\begin{figure}[!t]
    \centering
    \vspace{2mm}
    \includegraphics[width=0.45\textwidth]{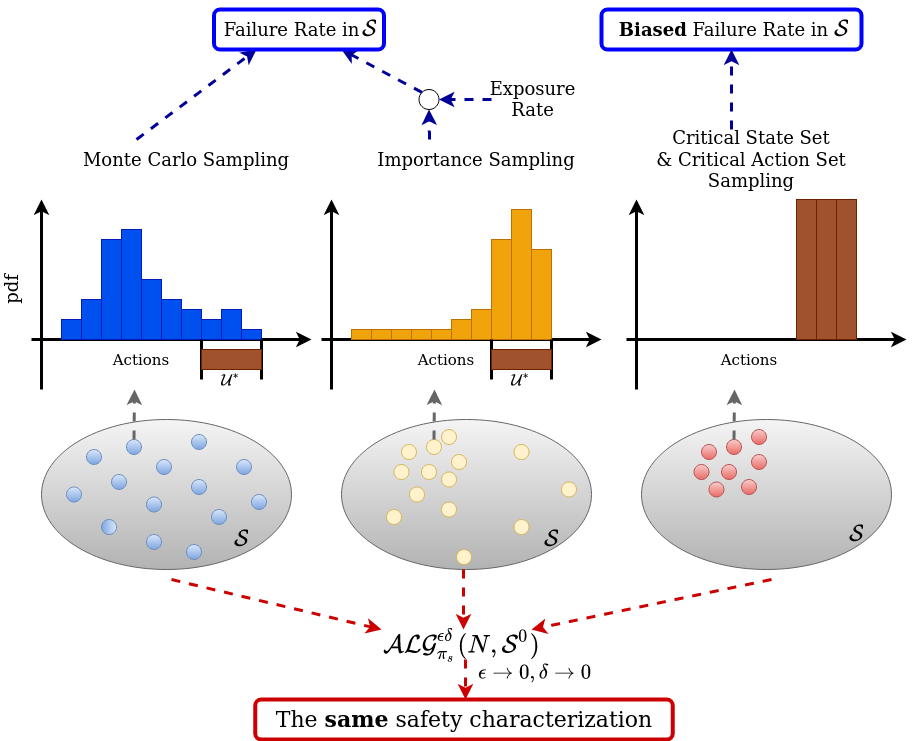}
    \caption{\small{Compare the safety characterization consensus between the traditional failure rate characterization and the proposed safe set characterization among various choices of scenario testing policies~$\pi_s(\cdot)$.}}
    \label{fig:pi_s}
    \vspace{-5mm}
\end{figure}

Consider the dominant characterization of failure rate in AV safety validation. It is well-known that the Monte Carlo sampling of states with naturalistic driving behavior model~\cite{yan2021distributionally} of $\pi_s(\cdot)$ and the various importance sampling based techniques~\cite{zhao2017accelerated, feng2021intelligent}, which have a biased focus towards high-risk state-actions, are all capable of approximating the unbiased estimate of failure rate. However, as one considers testing policies that are dedicated to safety-critical cases~\cite{corso2019adaptive,capito2020modeled}, many such ``worst-case-only" methodologies fail to reach the safety characterization consensus (as shown conceptually in Fig.~\ref{fig:pi_s}).

As one moves from the classic collision rate characterization to the proposed Problem~\ref{prob} and Algorithm~\ref{alg:quant}, the safety characterization consensus property is also strengthened. In particular, as long as the policy induces a sufficient coverage of a subset of ``critical" actions, all algorithm variants with different scenario testing policies converge to the same safe set asymptotically. This is formally presented as follows. Recall that $\mathcal{U}_s$ denotes the nominal action space for the controllable factors in the scenario. Let $\mathcal{U}= \mathcal{U}_s$. 
\begin{definition}{[\textbf{Feasible Action Set}]}\label{def:feasible-u-set}
    Consider the scenario testing policy $\pi_s(\cdot)$ as defined in~\eqref{eq:s-ctrl}. Given $\s\in\mathcal{S}$, let the feasible action space for $\pi_s(\cdot)$ and $\s$ be defined as:
    \begin{equation}\label{eq:feasible-u-set}
        \mathcal{U}(\s)=\{\uu \in \mathcal{U} \mid \exists \boldsymbol\omega_{\pi_s}\in\Omega_{\pi_s}, \pi_s(\s;\boldsymbol\omega_{\pi_s})=\uu \}.
    \end{equation}
\end{definition}
\begin{definition}{[\textbf{Critical Action Set}]}\label{def:b-u-set}
    Consider the scenario dynamics~\eqref{eq:scenario-dyn}. The critical action set for $\s\in \Sigma$ and $\Sigma\subseteq\mathcal{S}$ is defined as 
    \begin{equation}
        \mathcal{U}'(\s,\Sigma) = \Big\{ \uu\in\mathcal{U} \mid \exists \hat{\boldsymbol\omega}\in\hat{\Omega}, \text{ s.t. }\hat{f}(\s,\uu;\hat{\boldsymbol\omega}) \notin \Sigma \Big\}.
    \end{equation}
\end{definition}
That is, the feasible action set includes all actions that are ``reachable" from $\s$. It is also obviously dependent upon the policy $\pi_s(\cdot)$. 
On the other hand, the critical action set specifies a set of all actions in $\mathcal{U}$ that could drive the next-step state to reach the exterior of the given set $\Sigma$ from $\s$, hence it is dependent upon the instantaneous state-set pair. As a result, we have the \emph{critical feasible action set} $\mathcal{U}^*(\s,\Sigma)=\mathcal{U}'(\s,\Sigma) \cap \mathcal{U}(\s)$.
We are now ready to present the asymptotic safety characterization consensus property as follows.
\begin{theorem}{[\textbf{Asymptotic Safety Characterization Consensus}]}\label{thm:consensus}
    Let $\hat{\pi}_s(\cdot)$ be the nominal testing policy with the critical feasible action set $\hat{\mathcal{U}}^*(\s,\Sigma)$ for the state-set pair $(\s, \Sigma) (\s\in\Sigma)$. Let $\hat{\Phi}_{\delta}^{\partial\Sigma_{\bar{s}}}$ be the $\epsilon\delta$-almost safe set obtained with probability at least $1-\beta$ for $\hat{\pi}_s(\cdot)$.
    Consider an arbitrary scenario testing policy $\pi_s(\cdot)$ in the form of~\eqref{eq:s-ctrl} with the critical feasible action set $\mathcal{U}^*(\s,\Sigma)$ and the $\epsilon\delta$-almost safe set $\Phi_{\delta}^{\partial\Sigma_{\bar{s}}}$ obtained with probability at least $1-\beta$ for $\pi_s(\cdot)$. 
    We have
    \begin{equation}
        \lim_{\epsilon\rightarrow0, \beta\rightarrow0, \boldsymbol\delta\rightarrow0}\ \hat{\Phi}_{\delta}^{\partial\Sigma_{\bar{s}}} = \lim_{\epsilon\rightarrow0, \beta\rightarrow0, \boldsymbol\delta\rightarrow0}\ \Phi_{\delta}^{\partial\Sigma_{\bar{s}}},
    \end{equation}
    if for all $\s \in \Phi_{\delta}^{\partial\Sigma_{\bar{s}}}$ sampled throughout the execution of Algorithm~\ref{alg:quant}, 
    \begin{equation}\label{eq:boudary_in_feasible}
        \hat{\mathcal{U}}^*(\s,\Phi_{\delta}^{\partial\Sigma_{\bar{s}}})=\mathcal{U}^*(\s,\Phi_{\delta}^{\partial\Sigma_{\bar{s}}}).
    \end{equation}
\end{theorem}
The sketch of proof for the above theorem is as follows. By Theorem~\ref{thm:opt-alg}, Algorithm~\ref{alg:alg-overview} is asymptotically optimal, hence it suffices to show that the condition~\eqref{eq:boudary_in_feasible} remains valid for all state-set pairs during the execution of Algorithm~\ref{alg:quant}. This further ensures the uniqueness of the optimal set and can be proved through contradiction. Note that pruning and exploration are the only two operations that could make changes to the instantaneous set, and the critical action set is the only set of actions that could activate pruning and exploration. 
That is, as long as the set of critical feasible actions of the given policy is the same as the nominal critical feasible action set, the safety characterization algorithm converges to the same outcome asymptotically. In practice, it is not necessary to sample from the exact critical action set, as long as the conditions in Theorem~\ref{thm:consensus} are satisfied.


We conclude the discussion of the safety characterization consensus by emphasizing that a more ``dangerous" $\pi_s(\cdot)$ satisfying Theorem~\ref{thm:consensus} will also take fewer samples for convergence. This is intuitive given that the risk biased policy ``prunes" the unsafe states from the instantaneous set more frequently. The case study in the next section also provides empirical evidence to support this point. However, the design of the testing policy $\pi_s(\cdot)$ is essentially case-specific. Without knowing the explicit formulation of $\pi_s(\cdot)$, it is difficult to identify how sampling efficiency can be improved rigorously. In theory, regardless of the design of $\pi_s(\cdot)$ (as long as Theorem~\ref{thm:consensus} is satisfied), Algorithm{~\ref{alg:quant}} has the following property.
\begin{remark}\label{rmk:complexity}
    The computational complexity of Algorithm{~\ref{alg:quant}} is {$O(N\log{N})$} with the brute-force algorithm's complexity being {$O(N)$}. Without the critical state set $\partial\mathcal{S}_{\bar{s}}$ (i.e., the algorithm explores all states in $\mathcal{S}$), the complexity becomes $O(N^2)$. Moreover, as $N$ is determined by the number of vertices on the graph, it essentially grows exponentially as the state dimension increases. In practice, for high-dimensional problems, one can relax the problem with appropriate choices of {$\delta$} and {$\epsilon$}.
\end{remark}
Although the proposed method still suffers from the curse-of-dimensionality in theory, the presented computational complexity is better than some of the model-based solution in approximating the backward reachable set~\cite{mitchell2004demonstrating}, which is conceptually similar to the $\epsilon\delta$-almost safe set studied by this letter. A more effective algorithm design is of future interest.

\section{Case Study: Control Barrier Function based Mobile Robot Collision Avoidance System}\label{sec:case}
In this section, we consider a mobile robot system based on the Control Barrier Function (CBF)~\cite{wang2017safety} for collision avoidance. The CBF method has been explored extensively in various robotic applications~\cite{ames2016control, wang2017safety}. More importantly, the method naturally comes with the notion of a forward invariant set to ensure the robot's safety in theory, which aligns with our notion of safety by Definition~\ref{def:safety}. This is a very unique property as very few methods have the capability to derive the explicit operable domain (i.e., very few methods know exactly where the subject is safe in expectation). This inspires our exploration in characterizing the discrepancies, if applicable, between the theory-driven safe set and the data-driven safe set in practice.

\subsection{Collision avoidance with control barrier function}
The following CBF construction is mostly adapted from~\cite{wang2017safety}. Consider the linear double integrator dynamics as 
\begin{equation}\label{eq:di-dyn}
    \dot{\x}=\begin{bmatrix}\dot{\p} \\ \dot{\vv}\end{bmatrix} = \begin{bmatrix}\vv \\ \acc\end{bmatrix},
\end{equation}
with position states $\p=[p_x, p_y]\in\R^2$, velocity states $\vv=[\vv_x, \vv_y]\in\R^2$, and acceleration control actions $\acc = [a_x, a_y] \in \mathcal{A}\subseteq \R^2$. Note that the above dynamics take a general control-affine system formulation as $\dot{\x}=f(\x)+g(\x)\acc$ for $f(\x)$ being zero and $g(\x)=\begin{bmatrix}0 &0 &1 & 1\end{bmatrix}$.
Consider a pair of robots admitting the same motion dynamics of~\eqref{eq:di-dyn}, denoted with the subscript 0 and 1 (e.g., $\p_0$ and $\p_1$ for the position states). Consider the following set defined by a level-set function $h: \R^8 \rightarrow \R_{\geq0}$:
\begin{equation}\label{eq:di-set}
    \Psi_{01}\!=\!\Big\{\!\x_{01}\!=\!\begin{bmatrix}\x_0 & \x_1\end{bmatrix}\!\mid\! h(\x_{01}):=h(\x_0, \x_1)\!\geq\!0\!\Big\}
\end{equation}
Intuitively, $\Psi_{01}$ induces the set of states within which one expects both robots to operate without collisions. Furthermore, the function $h$ is referred to as a zeroing control barrier function (ZCBF) if there exists a locally Lipschitz extended class $\mathcal{K}$ function $\alpha(\cdot)$ such that
\begin{equation}\label{eq:zcbf}
    L_f h(\x_{01})\!+\!L_g h(\x_{01})\dot{\x}_{01}\!+\!\alpha(h(\x_{01}))\!\geq\!0,\forall \x_{01}\!\in\!\Psi_{01}.
\end{equation} 
By~\eqref{eq:di-dyn}, the coupled control action $\acc_{0}$, $\acc_1$ can be obtained by applying a linear projection of $\dot{\x}_{01}$, and the essential form of~\eqref{eq:zcbf} induces a set of linear inequalities associated with the coupled control of $\acc_0$ and $\acc_1$ in a centralized manner. By the analysis results of the CBF method, if the actions, $\acc_0(t)$ and $\acc_1(t)$, satisfy~\eqref{eq:zcbf} for all $t\geq0$ and for all $\x_{01}(0) \in \Psi_{01}$, then $\x_{01}(t) \in \Psi_{01}, \forall t \in\R_{>0}$. 
This further implies the forward set invariance similar to our previous Definition~\ref{def:safety}. That is, the CBF-based collision avoidance algorithm naturally induces a forward invariant set as the safe operable domain of the robot. This can be treated as the original ODD of the test subject. However, note that the derivation of the CBF-based methodology is essentially done in continuous-time, but it is executed at discrete time steps in practice. Moreover, the high-level planner is not considered in the original CBF design. Both factors could cause discrepancies between the expected performance by theory and the actual performance in scenario sampling. 

\subsection{Safety characterization}

\begin{figure*}
    \centering
    \vspace{2mm}
      \centering
      \includegraphics[trim={1cm 0.5cm 1cm 1cm},clip,width=0.99\linewidth]{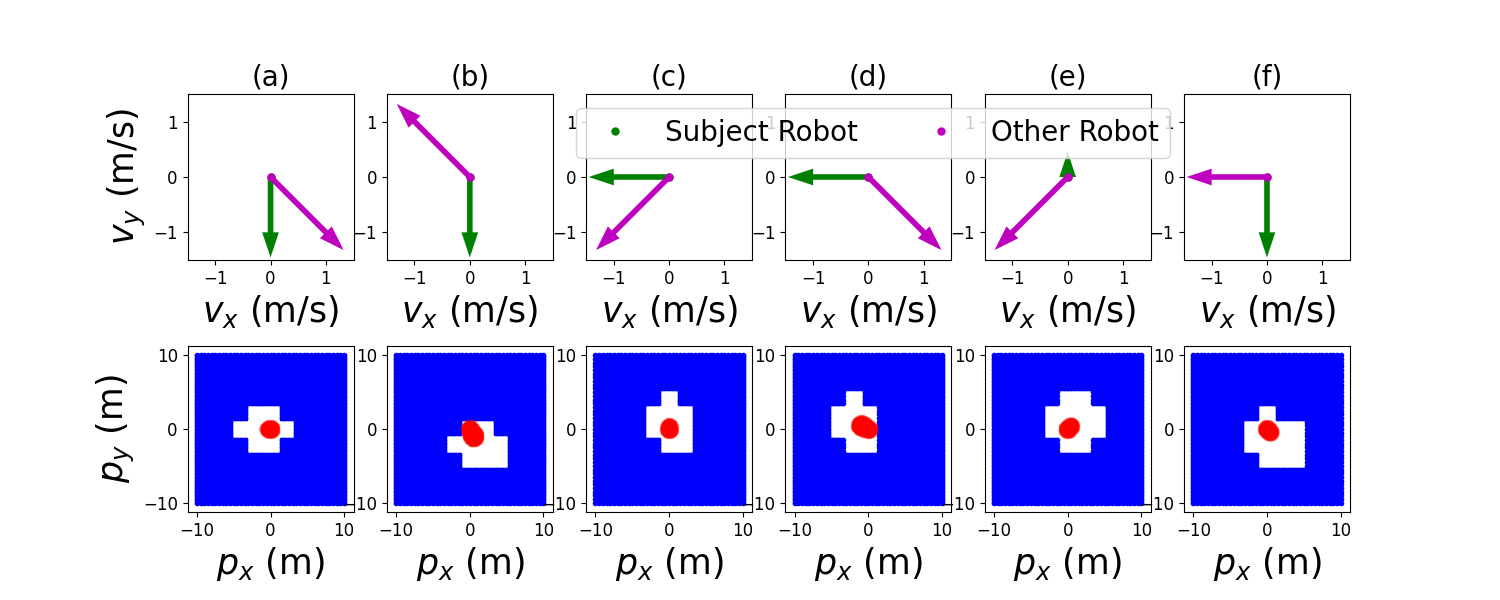}
      \caption{\small{Some selected subsets of the characterized $\epsilon\delta$-almost safe set ($\epsilon=0.001$, $\boldsymbol\delta=[0.5\ 0.5\ 0.2\ 0.2\ 0.2\ 0.2]$) with a confidence level of at least $0.9$ and the scenario policy of $\pi_s^{\text{CBF}}(\cdot)$. The first-row subplots illustrate the velocity for both robots. Given the velocity pair, the second-row subplots induce the theoretical safe operable set (complement of the red color region $\mathcal{S}\setminus\Psi_{01}$) and the characterized safe set (blue region) in practice.}}
      \label{fig:cbf_slices}
      \vspace{-2mm}
\end{figure*}

\begin{figure}
\centering
  \includegraphics[trim={0cm 0 0cm 0.5cm},clip,width=0.9\linewidth]{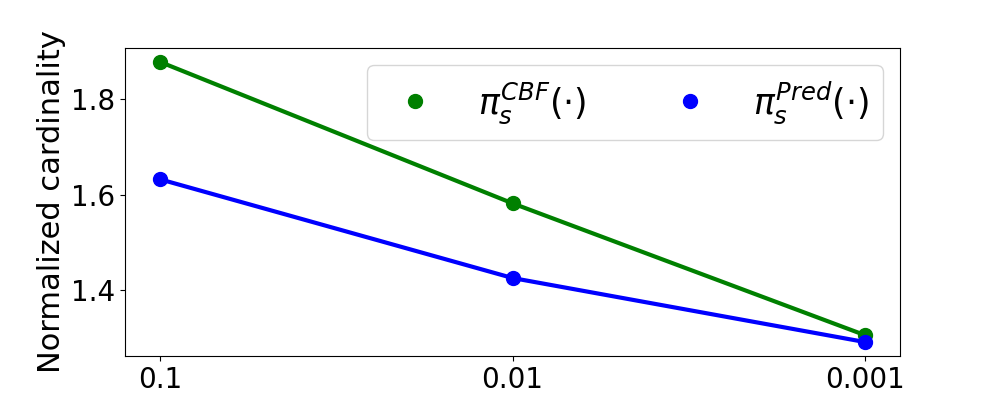}
  \caption{\small{The cardinality of the characterized $\epsilon\delta$-almost safe set with different choices of $\epsilon$. Each cardinality value is obtained over 20 trials of Algorithm~\ref{alg:quant} over the same group of 20 random seeds.}}
  \label{fig:cbf_epsilon}
  \vspace{-5mm}
\end{figure}

\begin{figure}[t]
  \centering
  \vspace{2mm}
  \includegraphics[trim={0cm 0 0cm 1cm},clip,width=0.9\linewidth]{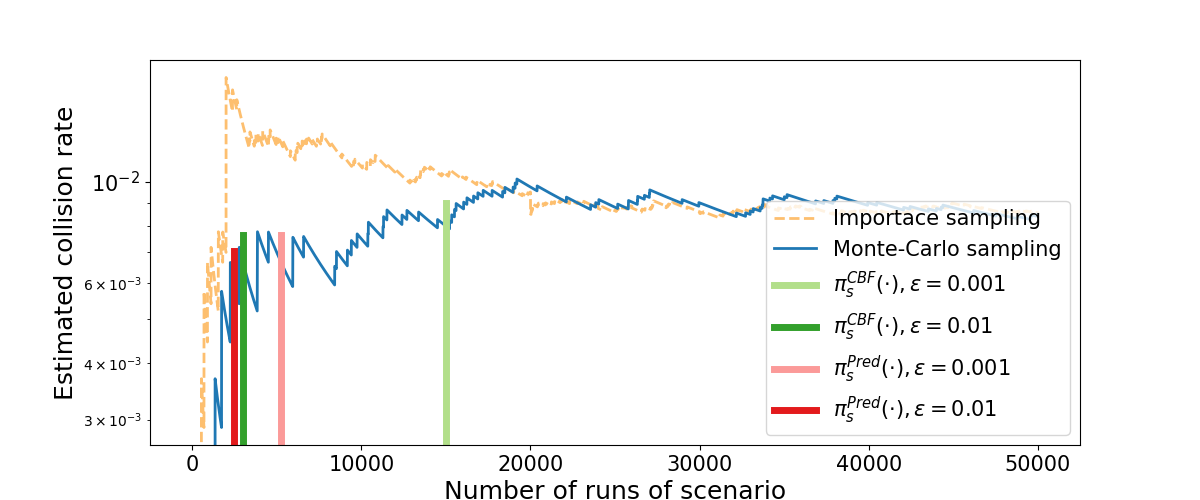}
  \caption{\small{An empirical sample efficiency comparison among the importance sampling based failure rate estimation (dashed orange line), the} Monte Carlo sampling of failure rate characterization (continuous blue line) and the proposed safety characterization scheme of different configurations (various vertical colored bars).}
  \label{fig:cbf_sample_efficiency}
  \vspace{-5mm}
\end{figure}

Consider the scenario system of two CBF-driven robots driven by the dynamics~\eqref{eq:di-dyn} and the CBF design the same as~\cite{wang2017safety}. We have the 6-dimensional state $\s = \begin{bmatrix}p_x & p_y & v_x^0 & v_y^0 & v_x^1 & v_y^1 \end{bmatrix} \in \mathcal{S} \subseteq [-10,10]^2\times[-1,1]^4$ with the distance offsets $p_x$ and $p_y$ between two robots on the planner domain (subject robot is at the origin), and the global velocity of both robots as $v_x^0$, $v_y^0$, $v_x^1$, $v_y^1$. Note that sufficiently large distance between robots will break the forward invariant condition by Definition~\ref{def:safety}, but is not of practical value for safety considerations. Hence the collected states for the runs of scenarios are clipped by the maximum positioning magnitude (10 m) as specified above. Although the velocity state is unbounded in theory, we have customized braking reference motion to ensure the satisfaction of the given bound in practice. This leaves the ``collision" as the only failure event of interest. Consider the circular robot with safety radius $0.5$m, hence the failure set $\mathcal{C}=\{\s\in\mathcal{S}\mid \norm{[p_x, p_y]}\leq 1\}$. Note that $\mathcal{S}\setminus\Psi_{01}$ is not necessarily the same as $\mathcal{C}$ (see the red zone in Fig.~\ref{fig:cbf_slices}). Let the acceleration controls for each robot satisfy $a_x^i\in[-1,1], a_y^i\in[-1,1], \forall i\in\{0,1\}$. For the subject robot (subscript 0), a target position is uniformly sampled within the admissible state space and remains persistent throughout the run of a scenario. The robot first propagates a goal-to-goal reference action based on the current position and the target position, and then calculates the nearest feasible action within the CBF induced safe admissible action space by solving a convex Quadratic Programming (QP) problem with linear constraints. This is the same policy as presented in~\cite{wang2017safety}.

For the other robot control (i.e., the scenario control), we consider two different policy constructions. The nominal policy $\pi_s^{\text{CBF}}(\cdot)$ executes the same policy as the subject robot control including the goal-to-goal planner and the CBF constrained QP. The predictive policy $\pi_s^{\text{Pred}}(\cdot)$ executes a predictive capturing strategy adapted from~\cite{capito2020modeled}. Instead of tracking the goal-to-goal reference, $\pi_s^{\text{Pred}}(\cdot)$ ``predicts" the future trajectory of the subject robot and seeks to track the predictive reference while admitting the CBF induced action constraints. It is immediate that $\pi_s^{\text{Pred}}(\cdot)$ implies higher risk than $\pi_s^{\text{CBF}}(\cdot)$. Furthermore, we also assign the predictive motion with (i) randomly selected look-ahead time horizon in $[0.2, 5]$ seconds, (ii) added Gaussian noise near the nominal predictive capturing policy, and (iii) action constraints induced by the same CBF condition with $\pi_s^{\text{CBF}}(\cdot)$. Hence, the condition of Theorem~\ref{thm:consensus} is satisfied (both policies share the same feasible action set induced by the CBF action constraints). In the practical execution of Algorithm~\ref{alg:alg-overview} and Algorithm~\ref{alg:quant}, we initialize the algorithms using $\mathcal{S}^0=\mathcal{S}\setminus \mathcal{C}$ and $\bar{\mathcal{S}}^0=\emptyset$. The time step for the discrete-time execution is $0.1$ second. The run of a scenario has a maximum time duration of 10 seconds. $\boldsymbol\delta$ decays from $[5\ 5\ 1\ 1\ 1\ 1]$ to $[0.5\ 0.5\ 0.2\ 0.2\ 0.2\ 0.2]$. The minimum $\bar{s}$ is $0.2$ in theory, we set $\bar{s}=0.5$ throughout the execution. Unless specified otherwise, $\beta=0.1$ (confidence level of $0.9$). 

The experimental results are presented in Fig.~\ref{fig:cbf_slices}, Fig.~\ref{fig:cbf_epsilon}, and Fig.~\ref{fig:cbf_sample_efficiency}. We further emphasize the following observations. 
In Fig.~\ref{fig:cbf_slices}, the white region shows the difference between the theoretical safe set $\Psi_{01}$ and the actual safe set under the given configurations. The instantaneous velocity states play an important role in causing such discrepancies. For example, consider the second column (b) in Fig.~\ref{fig:cbf_slices}. If the other robot is located at the upper left side of the subject robot, they are almost heading towards opposite directions, hence the boundary of $\Psi_{01}$ (upper left boundary of the red zone) aligns with the characterized safe set boundary. On the other hand, if the other robot is located at the lower right side, the fact that the robots are heading towards each other and the CBF-based control is executed at discrete-time, results in an extra gap required to ensure safety.
In Fig.~\ref{fig:cbf_epsilon}, as $\epsilon$ decreases, the average cardinality for the characterized safe set also decreases for both scenario testing policies. Moreover, $\pi_s^{\text{Pred}}(\cdot)$ tends to give a more accurate estimate of the safe set (with smaller cardinality) for the same value of $\epsilon$. However, as $\epsilon=0.001$, the two policies tend to achieve the same approximated cardinality value, which aligns with the theoretical analysis provided by Theorem~\ref{thm:consensus}.
Finally, in Fig.~\ref{fig:cbf_sample_efficiency}, within the group of 4 different characterizations, smaller $\epsilon$ with the predictive capturing scenario policy requires fewer samples to converge. Moreover, let $|\Phi_{\delta}^{0.001}|$ be the averaged approximated cardinality of the $\epsilon\delta$-almost safe sets obtained from $\pi_s^{\text{Pred}}(\cdot)$ and $\epsilon=0.001$ using $5250$ runs of the scenario on average. We have $1-\frac{|\Phi_{\delta}^{0.001}|}{|\mathcal{S}\setminus \mathcal{C}|}\approx0.01078$. This is very close to the collision rate estimate obtained from the Monte Carlo sampling approach ($0.01089$ with $50000$ samples), but the collision rate estimate requires nearly $10$ times more runs of the scenario. The observation also empirically validates the analysis in Remark~\ref{rmk:equal_to_fr}. A similar observation also generalizes to the comparison against the importance sampling based failure rate estimate adapted from~\cite{zhao2017accelerated}. Note that the pre-collected data for initializing the importance function estimate is not counted in the runs of scenarios illustrated in the figure. As a result, within the particular case being studied, most variants of the proposed method exhibit better sampling efficiency than other sampling-based failure rate estimate techniques.



\section{Conclusion} \label{sec:conclusion}
In this letter, we have presented a novel black-box system safety characterization criterion, an asymptotically optimal algorithm that solves the proposed optimal safety characterization problem, and an asymptotic safety characterization consensus property of the proposed algorithm. It is of future interest to explore the method with more complex problems of practical interest. That will primarily rely on a more case-specific design of testing policy $\pi_s(\cdot)$ that satisfies Theorem~\ref{thm:consensus}. Another possible line of research is to expand the applicability of the method to non-scenario-based testing regime such as the on-road test~\cite{kalra2016driving}.
\vspace{-3mm}
\bibliographystyle{IEEEtran}
\bibliography{mybibfile}

\end{document}